\documentclass{article} 
\usepackage{arxiv,times}

\usepackage{amsmath,amsfonts,bm}

\def\eqref#1{equation~\ref{#1}}

\def\1{\bm{1}}

\DeclareMathAlphabet{\mathsfit}{\encodingdefault}{\sfdefault}{m}{sl}
\SetMathAlphabet{\mathsfit}{bold}{\encodingdefault}{\sfdefault}{bx}{n}

\usepackage{hyperref}
\usepackage{url}
\usepackage{hyperref}       
\usepackage{url}            
\usepackage{booktabs}       
\usepackage{amsfonts}       
\usepackage{amssymb}        
\usepackage{nicefrac}      
\usepackage[table,dvipsnames]{xcolor}         
\usepackage{tabularx}       
\usepackage{multirow}       
\usepackage{array}          
\usepackage{makecell}       
\usepackage{adjustbox}      
\usepackage{threeparttable} 
\usepackage{amsmath}        
\usepackage{caption}        
\usepackage{pgf}            
\usepackage{graphicx}       
\usepackage{import}         

\usepackage[capitalize]{cleveref}
\usepackage{comment}
\usepackage{csquotes}
\usepackage{tabularx}
\usepackage{siunitx}
\usepackage{makecell}
\usepackage[most]{tcolorbox}

\newtcolorbox{promptbox}[1][]{
    colback=black!1!white,      
    colframe=black!2!white,    
    fonttitle=\bfseries,
    coltitle=black,
    title=#1,
    breakable, 
    left=6mm,
    enhanced,
    attach boxed title to top left={yshift=-2mm, xshift=4mm},
    boxed title style={
        colback=black!5!white, 
    }
}

\newcommand{\mypara}[1]{\textbf{#1.}}

\iclrfinalcopy

\title{Reference-Free Rating of LLM Responses via Latent Information}

\author{
\hspace{-13.25pt}
\begin{tabular}{llllll}
\multicolumn{2}{l}{Leander Girrbach\textsuperscript{\normalfont 1,3}} & 
\multicolumn{2}{l}{Chi-Ping Su\textsuperscript{\normalfont 2}} &
\multicolumn{2}{l}{Tankred Saanum\textsuperscript{\normalfont 4}} \\
\multicolumn{2}{l}{Richard Socher\textsuperscript{\normalfont 5}} &
\multicolumn{2}{l}{Eric Schulz\textsuperscript{\normalfont 3}} &
\multicolumn{2}{l}{Zeynep Akata\textsuperscript{\normalfont 1,3}} \\
\phantom{\rule{0.15\linewidth}{3pt}} & 
\phantom{\rule{0.15\linewidth}{3pt}} &
\phantom{\rule{0.15\linewidth}{3pt}} &
\phantom{\rule{0.15\linewidth}{3pt}} &
\phantom{\rule{0.15\linewidth}{3pt}} &
\phantom{\rule{0.15\linewidth}{3pt}} \\
\multicolumn{6}{l}{\normalfont{\textsuperscript{1}Technical University of Munich,\;Munich Center for Machine Learning,\; MDSI}} \\
\multicolumn{3}{l}{\normalfont{\textsuperscript{2}National Yang Ming Chiao Tung University}} &
\multicolumn{3}{l}{\normalfont{\textsuperscript{3}Helmholtz Munich}} \\
\multicolumn{3}{l}{\normalfont{\textsuperscript{4}Harvard University}} &
\multicolumn{3}{l}{\normalfont{\textsuperscript{5}you.com}}
\end{tabular}
}

\begin{document}

\maketitle

\begin{abstract}
How reliable are single-response LLM-as-a-judge ratings without references, and can we obtain fine-grained, deterministic scores in this setting? We study the common practice of asking a judge model to assign Likert-scale scores to free-text responses and show two systematic issues: scores are unstable under sampling and poorly calibrated, leading to compression near the top of the scale and frequent ties. We then propose and evaluate \emph{Latent Judges}, which derive scalar ratings from internal model signals: (i) probability-weighted scores over integer ratings, (ii) verifier-style probabilities of \enquote{yes}, and (iii) linear probes trained on model activations at the rating position. Across a broad suite of pairwise and single-rating benchmarks, latent methods match or surpass standard prompting, with consistent gains on pairwise accuracy and listwise ranking relevant to Best-of-$N$ selection. Probability-weighted scores achieve the strongest single-rating correlations, while probes recover useful signals when output logits are miscalibrated. These results indicate that latent information provides deterministic and more discriminative signals for reference-free evaluation, and can improve selection and training approaches like Best-of-$N$, multi-teacher distillation, and routing.
\end{abstract}

\section{Introduction} 
\label{sec:intro}

Different Large Language Models (LLMs) have distinct strengths and weaknesses, and even a single model can produce responses of varying quality to the same prompt. This variability has been productively exploited by Best-of-$N$ sampling at inference \citep{cobbe2021training,zhang2025generative} and by post-training with Group Relative Policy Optimization (GRPO) \citep{guo2025deepseek}. It also enables routing, which selects the right model for a given input \citep{ong2025routellm,zhang2025avengers}, and multi-teacher distillation, which transfers knowledge across models \citep{timiryasov2023baby,roth2024fantastic,tian2025beyond,gu2025capturing}. Across these settings, we often need \emph{reference-free} judgements of response quality that are \emph{fine-grained} and, ideally, \emph{deterministic}.

Much of the prior work has focused on \emph{verifiable} settings, such as code, math, factuality, or formatting, where correctness can be checked objectively \citep{guo2025deepseek,zhang2025generative}. In contrast, evaluating the quality of responses to arbitrary prompts is harder \citep{gehrmann2021gem}. Here, the prevailing approach is the LLM-as-a-Judge paradigm \citep{zheng2023judging}, in which a model rates (or compares) responses, typically on a 5-point Likert scale \citep{lambert2024tulu,hashemi2024llm,lee2024checkeval}.

However, our analysis reveals two important issues when obtaining \emph{single-response}, \emph{reference-free} ratings via prompting. First, unless decoding greedily, scores are \emph{unstable}: the same response can receive different ratings across runs. Second, ratings are often \emph{poorly calibrated}: they compress near the top of the scale, leading to frequent ties and limited discriminability. These problems arise from generating discrete tokens on bounded scales under stochastic decoding, and they persist even with strong judge models.

These limitations matter in practice. Reward learning benefits from \emph{fine-grained, unconstrained} scalar signals that better reflect true quality and are more robust to distractors \citep{tripathi2025pairwise}. Best-of-$N$ selection \citep{cobbe2021training,lightman2023let} and multi-teacher distillation require clear, tie-resistant rankings \emph{without} reference answers for calibration. Likewise, routing needs consistent per-response scores to choose among models. In short, many high-impact applications require \emph{deterministic and discriminative} reference-free evaluation.

To address this, we propose and evaluate \textit{Latent Judges}, which derive scalar ratings from internal model signals instead of only from generated tokens. We study three complementary families: (i) \emph{probability-weighted ratings}, which compute the expectation over integer scores using the model’s next-token distribution; (ii) \emph{verifier-style ratings}, which use the probability of \enquote{yes} in a binary \enquote{is this response good?} query; and (iii) \emph{latent probes}, lightweight classifiers trained on hidden activations at the rating position. These methods expose \emph{latent information} that is inherently real-valued and deterministic (for fixed inputs), can be rescaled to address calibration, and can recover useful quality signals even when output logits are miscalibrated.

In summary, our contributions are: (1) We systematically evaluate weaknesses of ordinal, single-response LLM-as-a-judge ratings, i.e.\ instability, compression, and ties, across a wide range of general and finetuned judge models. 
(2) We propose to mitigate these limitations by using \emph{latent} model information (probabilities and probes) to produce deterministic, fine-grained ratings. 
(3) We demonstrate that across standard LLM-as-a-judge metrics, latent judges perform on par with or better than prompting baselines. 
(4) We show how these insights improve practically relevant applications, including listwise ranking for Best-of-$N$ selection and the design of LLM routers.

\section{Related Work}
\label{sec:related}

\mypara{LLM-as-a-Judge}
LLM-as-a-Judge has emerged as a practical alternative to traditional human and metric-based evaluation, with models such as GPT-4 shown to align closely with human judgments \citep{zheng2023judging,li2024llms}. This paradigm has been applied to holistic quality scoring \citep{kim2024prometheus}, pairwise preference comparisons \citep{zheng2023judging}, and multi-rubric assessments \citep{lee2024checkeval,hashemi2024llm}, as well as domain-specific tasks like medical text generation \citep{brake2024comparing}, legal reasoning \citep{ryu2023retrieval}, and financial analysis \citep{xie2023pixiu}.

Recent work has sought to improve the reliability and faithfulness of evaluation through \textit{prompt-based methods} (e.g., GPTScore \citep{fu2024gptscore}, G-Eval \citep{liu2023g}), which guide evaluation with optimized prompts, explicit rubrics, or reasoning. \textit{Fine-tuned judges} such as JudgeLM \citep{zhu2025judgelm}, Auto-J \citep{li2023generative}, Prometheus \citep{kim2024prometheus,kim2024prometheus2}, and Themis \citep{hu2025training} directly adapt models to human preferences. However, this requires specialized data collection and model training and may not generalize well beyond the training set \citep{huang2024empirical}.

Finally, adapting LLM-as-a-judge to settings beyond pairwise comparison, which is relevant in practice,  is challenging. This is because of the inherent limitations of Likert scale ratings, a point we demonstrate extensively in this paper. Moreover, pairwise comparisons scale poorly to listwise rankings due to context limits, order bias, and non-transitivity \citep{xu2025investigating,hu2025language}.

\mypara{Verifiers}
To evaluate correctness in domains where objective correctness of responses can be determined, like math reasoning or code generation, training verifiers is successful \citep{cobbe2021training,uesato2022solving,wang2023math,lightman2023let,yu2024ovm,luo2024improve,hosseini2024vstar}. Verifiers classify responses as correct or incorrect, which serves as a reward model and helps Best-of-$N$ selection \citep{cobbe2021training,zhang2025generative}. Beyond trained verifiers, having a binary classifier assess the degree of correctness via the probability of \enquote{yes} has found application in math/coding \citep{zhang2025generative} and prompt following evaluation in text-to-image generation \citep{lin2024evaluating}. Our work extends this method to arbitrary natural language generation, where reference answers may not exist and evaluation requires going beyond binary correctness.
 
\mypara{Latent Probing}
LLMs encode rich information about the input text in their latent activations \citep{peters2018deep,devlin2019bert}. Many works have explored how to extract this knowledge using lightweight classifiers, i.e.\ \textit{probes} \citep{veldhoen2016diagnostic,ettinger2016probing,alain2017understanding,adi2017finegrained,conneau2018you}. These representations become even more informative when combined with the strong reasoning capabilities of LLMs through targeted prompting \citep{zou2023representation,marks2024geometry,yang2024large}, which enables applications such as steering \citep{turner2023steering,rimsky2024steering}, hallucination detection \citep{obeso2025real,orgad2025llms}, and detecting true or false statements \citep{marks2024geometry,maiya2025improving}.
Building on this, we explore probing as robust text evaluation. Specifically, we extract logits from judge LLMs and train probing classifiers on their internal activations to rate a response, which transforms these latent signals into stable and fine-grained ratings, overcoming the key shortcomings of existing LLM-as-a-Judge methods.

\section{Limitations of LLM-as-a-Judge and How to Fix Them}

\subsection{Uncovering Weaknesses of LLM-as-a-Judge Approaches}
\label{sec:method:weaknesses}

We systematically examine the limitations of using LLMs to assign holistic ratings to single responses without a reference. This setting is realistic and important, for example, for Best-of-$N$ selection and multi-teacher distillation, but our analysis reveals two weaknesses: ratings are inconsistent across runs and poorly calibrated. As a result, scores fluctuate with the decoding seed and concentrate near the top of the scale, limiting their ability to distinguish response quality.

\mypara{Data and Models}
Our experiments use the Tülu Preference Mixture dataset \citep{lambert2024tulu}, which contains 273{,}000 prompts paired with chosen and rejected responses. We sample 5{,}000 prompts and their responses to evaluate 12 judge models, including three trained specifically as judges: Prometheus-v2 8B \citep{kim2024prometheus2}, Selene-1 Mini 8B, and Selene-1 70B \citep{alexandru2025atla}.  Models are prompted with two variants of the Prometheus template \citep{kim2024prometheus,kim2024prometheus2}: one using a 1--5 scale and another using a 1--10 scale. The concrete prompts are shown in \cref{sec:supp:prompts:baselines}. Ratings are generated with temperature 0.7, with 10 scores per response sampled using different seeds. This setup allows us to assess variability and calibration.

\begin{figure}[t]
\centering
\includegraphics[width=\linewidth]{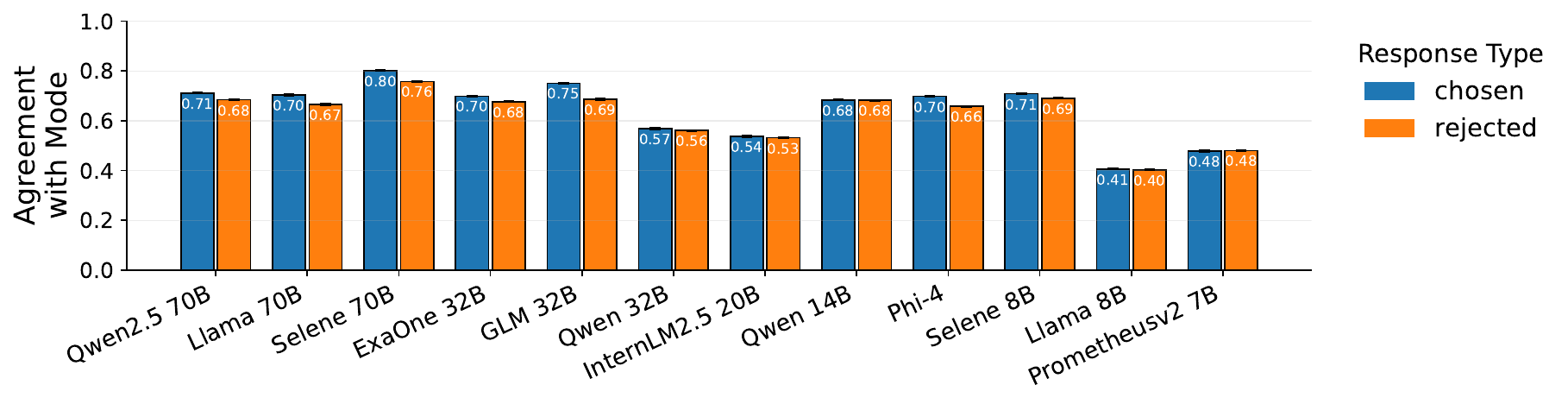}
\caption{Agreement of ratings with the mode across 10 sampled runs. Even the most consistent models fall below 80\% agreement, and some near 40\%.}
\label{fig:weaknesses:inconsistent}
\end{figure}

\mypara{Finding: LLM judges are inconsistent}
Figure~\ref{fig:weaknesses:inconsistent} reports the agreement of individual ratings with the mode across 10 samples. Even the most stable models, such as Qwen2.5~70B and Selene~70B, reach only 70--80\% agreement. Others, including Prometheus-v2~7B and Llama~8B, drop to 40--50\%. Thus, one in five ratings, and often more, diverges from the most frequent score. Some variation is expected under stochastic decoding, but this level undermines reliability. Because scores cluster in a narrow range, even small inconsistencies have a large impact.

\begin{figure}[t]
\centering
\includegraphics[width=\linewidth]{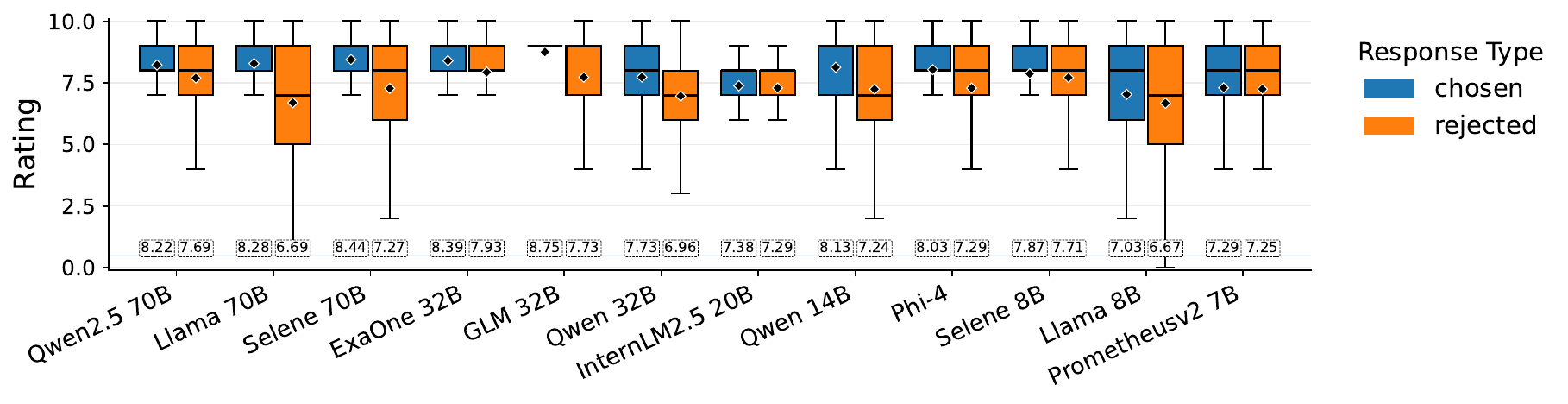}
\caption{Average ratings for chosen and rejected responses across judge models. Scores are high overall and differences are small.}
\label{fig:weaknesses:calibration}
\end{figure}

\mypara{Finding: LLM ratings are not calibrated}
Calibration is another issue. Figure~\ref{fig:weaknesses:calibration} shows that chosen responses nearly always receive high scores (often above 8 on a 1--10 scale), while rejected responses are only slightly lower. For instance, Qwen2.5~70B and Selene-1 assign averages between 7 and 8 even to rejected responses. This compression obscures meaningful differences: without a reference, models default to generous ratings because most responses seem strong in isolation.

High scores also lead to frequent ties. Table~\ref{tab:weaknesses:tuluagreement} compares strict and lenient agreement with GPT-4 preferences. Strict agreement, requiring chosen responses to score higher, is low across all models (e.g., 50.1\% for Llama~70B and 16.9\% for InternLM2.5~20B on the 1--5 scale). Lenient agreement, which counts ties as correct, is much higher (e.g., 92.6\% and 85.9\% respectively). The large gaps show that ties are pervasive and that single-response ratings lack discriminative power.

\begin{table}[t]
\centering
\setlength{\tabcolsep}{6pt}
\resizebox{\linewidth}{!}{
\begin{tabular}{clrrrrrrrrrrrr}
\toprule
& Model & \makecell[r]{Qwen2.5 \\ 70B} & \makecell[r]{Llama3 \\ 70B} & \makecell[r]{Selene \\ 70B} & \makecell[r]{ExaOne3.5 \\ 32B} & \makecell[r]{GLM4 \\ 32B} & \makecell[r]{Qwen3 \\ 32B} & \makecell[r]{Int.\ LM2.5 \\ 20B} & \makecell[r]{Qwen3 \\ 14B} & \makecell[r]{Phi-4 \\ 14B} & \makecell[r]{Selene \\ 8B} & \makecell[r]{Llama3 \\ 8B} & \makecell[r]{Prom.v2 \\ 7B} \\
\midrule
\multirow{2}{*}{5} & Strict   & 35.7 & 50.1 & 50.0 & 29.1 & 39.8 & 37.1 & 16.9 & 40.9 & 29.4 & 23.3 & 32.5 & 20.6 \\
& Lenient  & 87.5 & 92.6 & 93.8 & 86.2 & 91.3 & 88.5 & 85.9 & 88.0 & 91.6 & 82.2 & 75.3 & 80.9 \\
\midrule
\multirow{2}{*}{10} & Strict  & 38.9 & 61.2 & 56.4 & 32.9 & 44.6 & 49.8 & 34.3 & 49.9 & 42.7 & 30.2 & 42.6 & 34.6 \\
& Lenient & 83.1 & 88.3 & 90.4 & 80.6 & 88.2 & 77.7 & 68.1 & 80.8 & 82.4 & 75.2 & 65.5 & 65.3 \\
\bottomrule
\end{tabular}
}
\caption{Agreement of judge models with GPT-4 preferences. Strict metrics require chosen responses to score higher than rejected ones; lenient metrics also allow ties. Large gaps indicate poor discriminability of single-response ratings.}
\label{tab:weaknesses:tuluagreement}
\end{table}

\mypara{Summary}
Overall, the holistic scoring of single responses without references is unreliable. Ratings vary substantially across runs, with even the best models disagreeing one-quarter of the time. Scores are also inflated and compressed near the top of the scale, producing frequent ties that obscure differences. While pairwise prompts can mitigate this when both responses are shown together, our analysis highlights fundamental weaknesses of single-response holistic scoring, which is the focus of this work.
These findings motivate us to look for methods that mitigate these issues by providing deterministic and discriminative ratings.

\subsection{Using Scores Derived from Latent Information to Rate Responses}
\label{sec:method:raters}

As shown in \cref{sec:method:weaknesses}, LLM-as-a-judge for single responses has several drawbacks. The ratings are often unstable, saturated near the top of the scale, and lack discriminative power. These issues stem from using discrete categories, bounded scales, and stochastic decoding \citep{holtzman2020curious}.
We investigate to what extent extracting the latent knowledge of a Judge LLM immediately after it processes the prompt can mitigate these issues. This approach offers, in theory, clear advantages: it is deterministic because it does not rely on sampling the LLM's output, and it is discriminative because its scores are real-valued instead of integers on an ordinal scale. Furthermore, these real-valued scores can be scaled and shifted to solve potential calibration issues.

To extract latent knowledge, we investigate three different methods. In
\textit{probability-weighted ratings}, we calculate a weighted average of integer ratings. We prompt the LLM to rate a response on a scale, such as 1 to 10, and to output only the rating. Then, we take the token probabilities for the next predicted token, extract the probabilities that correspond to the integers on the scale, and compute a weighted average. This is formally expressed as:
\begin{equation}
S_p(\text{prompt}) = \sum_{i=1}^n n\times p_{\text{LLM}}(n\ {|\ \text{prompt})}
\end{equation}
In \textit{binary} or \textit{verifier-style ratings}, we ask the LLM if the given response is good for the given prompt, and the LLM should only output either \enquote{yes} or \enquote{no}. We then use the probability of predicting \enquote{yes} as the rating, formally:
\begin{equation}
S_b(\text{prompt}) = p_{\text{LLM}}(\text{yes}\ |\ \text{prompt})
\end{equation}
Finally, in \textit{latent probing}, we train linear probes on the latent activations of a Judge LLM. Specifically, we extract the residual stream activation at the position of the next predicted token after prompting the model to evaluate a response. This option is particularly useful when the LLM's logits are not well-calibrated and may not faithfully reflect its internal states. Formally, for a Judge LLM $f_\theta$, given an input sequence of length $T$, we extract activation $z^{(l)} \in \mathbb{R}^d$ from layer $l$ at position $T{+}1$. We then train a lightweight probe $g_\phi : \mathbb{R}^d \to \mathbb{R}$ that predicts a quality score $\hat{y} = g_\phi(z^{(l)})$. Probes can be linear or small MLPs. Details on how we train probes on activations are in \cref{sec:supp:probetraining}.

\section{Experimental Evaluation of LLM-as-a-Judge Benchmarks}

\subsection{Pairwise and Single-Rating Benchmarks}
We evaluate models on two types of benchmarks: \textit{Pairwise} and \textit{Single-Rating}. These are the traditional baselines to evaluate LLM Judges \citep{kim2024prometheus,kim2024prometheus2,alexandru2025atla}. In \textit{pairwise} benchmarks, we assess if the LLM Judge assigns a higher score to a response preferred by humans or GPT-4 than to its dispreferred counterpart. In \textit{single-rating} benchmarks, we evaluate if the scores correlate with ground-truth ratings from human raters or GPT-4.

\mypara{Pairwise Benchmarks}
We evaluate models on the following benchmarks: MT-Bench Human Preferences \citep{zheng2023judging}, RewardBench \citep{lambert2025rewardbench}, HHH-Alignment \citep{askell2021general}, RewardBench-2 \citep{malik2025rewardbench}, Auto-J \citep{li2023generative}, LFQA \citep{xu2023critical}, PreferenceBench \citep{kim2024prometheus}, and JudgeBench \citep{tan2025judgebench}. All benchmarks consist of $\langle \text{prompt, chosen, rejected} \rangle$ triplets. We only consider unambiguous preferences, where there is a chosen and a rejected response, and we exclude ties, as modeling ties is not the focus of this paper.

For MT-Bench and HHH-Alignment, we use deduplicated versions from \citep{kim2024prometheus}, removing all repeated $\langle \text{prompt, chosen, rejected} \rangle$ triplets. We also include 10,000-sample subsets from the Tülu Preference Mixture \citep{lambert2024tulu} and UltraFeedback \citep{cui2024ultrafeedback} as additional evaluation sets. The standard evaluation metric for these benchmarks is accuracy, which measures the rate at which predicted rankings agree with the ground truth. Benchmark sizes are in \cref{tab:benchmarks:pairwise:sizes}.

\begin{table}[t]
\centering
\small
\setlength{\tabcolsep}{8pt}
\begin{tabularx}{0.85\textwidth}{@{}l S[table-format=4] l S[table-format=4] l S[table-format=4]@{}}
\toprule
\textbf{Dataset} & {\textbf{Size}} & 
\textbf{Dataset} & {\textbf{Size}} &
\textbf{Dataset} & {\textbf{Size}} \\
\midrule
Auto-J          & 1019 &
JudgeBench      & 620  &
LFQA            & 1059 \\
PreferenceBench & 1998 &
RewardBench     & 2984 &
RewardBench-2   & 1825 \\
MT-Bench        & 890  &
HHH-Alignment   & 221  &
Tülu Mixture    & 10000 \\
UltraFeedback   & 10000 \\
\bottomrule
\end{tabularx}
\caption{Sizes of pairwise evaluation benchmarks.}
\label{tab:benchmarks:pairwise:sizes}
\end{table}

\mypara{Single-Rating Benchmarks} We include the following single-rating benchmarks: FLASK \citep{ye2024flask}, MTBench \citep{zheng2023judging}, Vicuna-Eval \citep{kim2024prometheus}, BiGGen Bench \citep{kim2025biggen}, and UltraFeedback \citep{cui2024ultrafeedback}. For FLASK and BiGGen Bench, both human and GPT-4 ratings are available. For UltraFeedback, we use only the rejected responses and their scores from the binarized version because the score distribution is skewed towards the max score.

Summary statistics and sizes are in \cref{tab:benchmarks:single:stats}. They vary in size, from 320 samples for Vicuna-Eval and MT-Bench to 68,805 for the GPT ratings in BiGGen Bench. All ratings are on a scale from 1 to 5, and the mean scores are positively skewed (i.e., above 3) in all benchmarks. However, the standard deviations indicate significant score variation, which is important for assessing qualitative differences between responses. Still, as all benchmarks resemble the 5-scale baseline and skew positive, this favors the baseline methods.

\begin{table}[t]
\centering
\small
\begin{tabular}{lrrrrrrr}
\toprule
  & \textbf{Vicuna-Eval} & \textbf{MT-Bench} & \textbf{UltraFeedback} & \multicolumn{2}{c}{\textbf{Flask}} & \multicolumn{2}{c}{\textbf{BiGGen Bench}} \\
 \cmidrule(lr){2-2} \cmidrule(lr){3-3} \cmidrule(lr){4-4} \cmidrule(lr){5-6} \cmidrule(lr){7-8}
  &  &  &  & Human & GPT & Human & GPT \\
\midrule
Size  & 320 & 320 & 60,917 & 2,001 & 2,001 & 2,780 & 68,805 \\
Mean  & 4.15 & 3.49 & 3.32 & 3.81 & 3.62 & 3.49 & 3.17 \\
Std   & 0.78 & 1.25 & 1.09 & 1.10 & 1.32 & 1.41 & 1.21 \\
\bottomrule
\end{tabular}%
\caption{Sizes and summary statistics of single-rating benchmarks.}
\label{tab:benchmarks:single:stats}
\end{table}

\subsection{LLM Models and Baselines}
We compare the rating methods described in \cref{sec:method:raters} to LLM Judges using the same 5-scale and 10-scale prompts evaluated in \cref{sec:method:weaknesses}, as this is our main point of comparison. Since this baseline produces frequent ties, as shown in \cref{sec:method:weaknesses}, we ensure a fair comparison by breaking ties randomly. For baselines, probability-weighted ratings, and verifier-style ratings, we use the following models: Phi-4 \citep{abdin2024phi}, Qwen3 14B and 32B \citep{yang2025qwen3}, Qwen2.5 70B, \citep{team2024qwen2}, Llama-3.3 70B \citep{dubey2024llama}, Prometheus-v2 7B \citep{kim2024prometheus2}, and Selene-1 70B \citep{alexandru2025atla}. We selected these models as the most capable judge models within our compute budget, and Selene-1 and Prometheus were selected as representatives of LLMs specifically fine-tuned for judging response quality. For latent probing, we only evaluate Qwen3 14B, Phi-4, and Prometheus v2 7B due to the increased cost of extracting embeddings.

\subsection{Results on Pairwise and Single-Rating Benchmarks}

Due to space constraints, we report results for Phi-4, Qwen3 14B, Prometheus, Llama 3.3 70B, and Selene-1, as well as only the 10-scale baseline. Results for other models and the 5-scale baseline confirm observations and are in \cref{sec:supp:additonal:results}.

\mypara{Pairwise Benchmarks}
\cref{tab:benchmarks:pairwise} reports accuracies across pairwise evaluation benchmarks. Four key observations stand out:
\textit{First}, probability-weighted and verifier-style scores consistently match or exceed the 10-scale baseline, with gains of up to 5 percentage points in average accuracy. This indicates that extracting probability distributions or binary logits yields more discriminative signals than discrete categorical outputs, while remaining competitive where the baseline performs well. \textit{Second}, specialized judge models such as Selene and Prometheus do not surpass general-purpose models (e.g., Llama, Phi-4). In fact, Prometheus fails entirely under probability-weighted and verifier setups, as it does not follow these prompts. This demonstrates that fine-tuning for judgment can compromise general model capabilities, reducing its applicability to alternative scoring schemes.

\textit{Third}, when raw logits are poorly calibrated (e.g., Qwen3 14B under weighted scoring, Prometheus across settings), latent probes recover useful internal signals. By training directly on hidden activations, probes extract stable and fine-grained information about response quality that is not accessible through the model’s output probabilities alone. This highlights latent probing as a general solution that can be applied to any judge model, regardless of its calibration. We also want to note that the latent probe can be further enhanced, to some degree (typically 1-2\%) by targeted hyperparameter tuning, but here we report results for the same parameter configurations.  \textit{Fourth}, the comparison is conservative with respect to alternative methods: for the 10-scale baseline, frequent ties are resolved by random breaking. Thus, a model that produces ties in half of all cases already reaches 65\% accuracy with only 40\% correct and 10\% incorrect predictions. The true discriminative ability of such baselines is therefore substantially lower than the reported numbers.

\paragraph{Single-Rating Benchmarks}
\cref{tab:benchmarks:single} reports Pearson correlations with ground-truth ratings, which themselves come from 5-point Likert scales or their averages. Here, we find that probability-weighted ratings and 10-scale ratings achieve the highest correlations, with strong models (Phi-4, Llama, Selene) reaching averages near 0.6. This suggests moderate to high correlation with human or GPT-4 ratings.  However, verifier-style scores perform significantly worse. Verifyer-style scores mostly concentrate close to 0 or 1, leading to consistently lower correlations. For latent probes, the BCE objective used for training likely also squashes outputs toward the extremes, reducing variance and degrading linear correlation. Combined with the discretization of ground truth, this limits their effectiveness on single-rating benchmarks.

\mypara{Summary}
Our extensive results on a large number of benchmarks show that approaches using internal judge representations serve as a valid replacement for traditional generative judges by addressing their limitations, such as non-determinism and saturated scales. Pairwise evaluation results show that their discriminativeness is superior or on par with the baselines. The linear correlation for single-rating evaluations is decent, especially for probability-weighted ratings. This is notable because the benchmarks closely resemble Likert scale ratings, which favors the 10-scale and 5-scale baselines. Finally, latent probing can recover model capabilities that are not accessible in training-free methods due to their miscalibrated outputs.

\begin{table*}[t]
\centering
\small
\setlength{\tabcolsep}{5pt}
\sisetup{
  table-number-alignment = center,
  table-format = 1.2,
  detect-weight = true,
  detect-inline-weight = math
}
\resizebox{\linewidth}{!}{
\begin{tabular}{
l
l
S S S S S S S S S S
@{\hspace{6pt}} | @{\hspace{6pt}}
S
}
\toprule
\multicolumn{2}{c}{\textbf{Setting}} & 
\multicolumn{10}{c}{\textbf{Benchmarks}} &
\multicolumn{1}{c}{\textbf{Average}} \\
\cmidrule(lr){1-2} \cmidrule(lr){3-12} \cmidrule(l){13-13}
 &  &
\multicolumn{1}{c}{Auto-J} &
\multicolumn{1}{c}{HHH} & 
\multicolumn{1}{c}{JB} &
\multicolumn{1}{c}{LFQA} &
\multicolumn{1}{c}{MTB} &
\multicolumn{1}{c}{PB} &
\multicolumn{1}{c}{RB-2} &
\multicolumn{1}{c}{RB} &
\multicolumn{1}{c}{Tülu} &
\multicolumn{1}{c}{UF} &
\multicolumn{1}{c}{\textbf{}} \\
\midrule
\addlinespace[2pt]
\multirow{5}{*}{\rotatebox{90}{Verifier}} 
 & Prometheus                       & 0.12 & 0.16 & 0.00 & 0.00 & 0.08 & 0.02 & 0.10 & 0.07 & 0.08 & 0.09 & 0.07 \\
 & Qwen3 14B                        & 0.76 & 0.85 & 0.69 & 0.72 & 0.64 & 0.85 & \textbf{0.88} & 0.87 & 0.74 & 0.78 & 0.78 \\
 & Phi-4                            & 0.72 & 0.88 & 0.67 & 0.59 & 0.64 & 0.87 & 0.86 & 0.87 & 0.74 & 0.77 & 0.76 \\
 & Llama 3.3 70B                 & 0.72 & 0.87 & 0.63 & 0.72 & 0.65 & 0.83 & 0.82 & 0.82 & 0.72 & 0.74 & 0.75 \\
 & Selene-1                         & 0.72 & 0.85 & 0.66 & 0.47 & 0.66 & 0.85 & 0.85 & 0.85 & 0.73 & 0.76 & 0.74 \\
\addlinespace[4pt]
\midrule
\addlinespace[2pt]
\multirow{5}{*}{\rotatebox{90}{Weighted}} 
 & Prometheus                      & 0.00 & 0.00 & 0.00 & 0.00 & 0.00 & 0.00 & 0.00 & 0.00 & 0.00 & 0.00 & 0.00 \\
 & Qwen3 14B                       & 0.66 & 0.83 & 0.43 & \textbf{0.77} & 0.57 & 0.84 & 0.58 & 0.59 & 0.54 & 0.61 & 0.64 \\
 & Phi-4                           & 0.77 & 0.89 & 0.69 & \textbf{0.77} & 0.67 & 0.92 & 0.85 & {0.89} & 0.75 & 0.82 & 0.80 \\
 & Llama 3.3 70B                & 0.79 & 0.91 & 0.65 & \textbf{0.77} & 0.67 & 0.91 & 0.85 & {0.89} & 0.75 & 0.83 & 0.80 \\
 & Selene-1 70B                    & 0.79 & \textbf{0.92} & 0.68 & 0.76 & 0.68 & 0.94 & 0.86 & \textbf{0.91} & 0.77 & \textbf{0.85} & \textbf{0.82} \\
\addlinespace[4pt]
\midrule
\addlinespace[2pt]
\multirow{5}{*}{\rotatebox{90}{10-Scale}} 
 & Prometheus                      & 0.67 & 0.71 & 0.54 & 0.64 & 0.60 & 0.89 & 0.51 & 0.71 & 0.62 & 0.63 & 0.66 \\
 & Qwen3 14B                       & 0.77 & 0.78 & 0.61 & 0.74 & 0.65 & 0.81 & 0.73 & 0.82 & 0.73 & 0.79 & 0.74 \\
 & Phi-4                           & 0.74 & 0.83 & 0.59 & 0.76 & 0.65 & 0.84 & 0.73 & 0.83 & 0.72 & 0.76 & 0.74 \\
 & Llama 3.3 70B               & 0.77 & 0.85 & 0.60 & 0.75 & 0.65 & 0.88 & 0.72 & 0.84 & 0.73 & 0.78 & 0.76 \\
 & Selene-1                        & 0.77 & 0.87 & 0.63 & 0.74 & 0.65 & 0.91 & 0.76 & 0.88 & 0.75 & 0.82 & 0.78 \\
\addlinespace[4pt]
\midrule
\addlinespace[2pt]
\multirow{6}{*}{\rotatebox{90}{Latent Probe}} 
 & Prometheus (Models)   & 0.75 & 0.75 & 0.80 & 0.62 & 0.60 & 0.93 & 0.77 & 0.68 & 0.85 & 0.57 &  0.73 \\
 & Prometheus (Tülu)     & 0.74 & 0.75 & 0.79 & 0.61 & 0.62 & \textbf{0.95} & 0.77 & 0.69 & 0.83 & 0.57 &  0.73 \\
 & Qwen3 14B (Models)    & 0.75 & 0.70 & 0.89 & 0.64 & \textbf{0.87} & 0.89 & 0.79 & 0.76 & 0.87 & 0.69 &  0.78 \\
 & Qwen3 14B (Tülu)      & 0.75 & 0.72 & \textbf{0.90} & 0.66 & 0.86 & 0.88 & 0.78 & 0.74 & 0.86 & 0.68 &  0.78 \\
 & Phi-4 (Models)        & 0.78 & 0.75 & 0.88 & 0.66 & 0.85 & 0.90 & 0.80 & 0.76 & 0.88 & 0.67 &  0.79 \\
 & Phi-4 (Tülu)          & \textbf{0.80} & 0.76 & 0.88 & 0.67 & 0.84 & 0.91 & 0.82 & 0.76 & \textbf{0.89} & 0.68 &  0.80 \\
\bottomrule
\end{tabular}
}
\caption{Pairwise evaluation accuracies on triplet datasets ($\langle$prompt, chosen, rejected$\rangle$): Auto-J, HHH-Alignment (HHH), JudgeBench (JB), LFQA, MT-Bench (MTB), PreferenceBench (PB), RewardBench-2 (RB-2), RewardBench (RB), Tülu Mixture (Tülu), and UltraFeedback (UF). Latent Probes require training; we denote the data source as Tülu = Preference pairs from \citep{lambert2024tulu} and Models = generated by strong /weak LLMs.
Best scores for each benchmark are in bold.}
\label{tab:benchmarks:pairwise}
\end{table*}

\begin{table*}[t]
\centering
\small
\setlength{\tabcolsep}{5pt}
\sisetup{
  table-number-alignment = center,
  table-format = 1.2,
  detect-weight = true,
  detect-inline-weight = math
}
\resizebox{\linewidth}{!}{
\begin{tabular}{
l
l
S S S S S S S
@{\hspace{6pt}} | @{\hspace{6pt}}
S
}
\toprule
\multicolumn{2}{c}{\textbf{Setting}} &
\multicolumn{7}{c}{\textbf{Benchmarks}} &
\multicolumn{1}{c}{\textbf{Average}} \\
\cmidrule(lr){1-2} \cmidrule(lr){3-9} \cmidrule(l){10-10}
 &  &
\multicolumn{1}{c}{BigGen-H} &
\multicolumn{1}{c}{BigGen-J} &
\multicolumn{1}{c}{Flask-G} &
\multicolumn{1}{c}{Flask-H} &
\multicolumn{1}{c}{MTB} &
\multicolumn{1}{c}{UF} &
\multicolumn{1}{c}{Vicuna} &
\multicolumn{1}{c}{\textbf{}} \\
\midrule
\addlinespace[2pt]
\multirow{5}{*}{\rotatebox{90}{Verifier}}
 & Prometheus                       & -0.00 & 0.00 & 0.06 & 0.07 & {\textemdash{}} & 0.03 & {\textemdash{}} &  0.03 \\
 & Qwen3 14B                        & 0.43 & 0.64 & 0.37 & 0.37 & 0.32 & 0.63 & 0.44 &  0.46 \\
 & Phi-4                            & 0.46 & 0.70 & 0.42 & 0.37 & 0.54 & 0.66 & 0.50 &  0.52 \\
 & Llama 3.3 70B                 & 0.39 & 0.66 & 0.44 & 0.35 & 0.56 & 0.70 & 0.60 &  0.53 \\
 & Selene-1                         & 0.42 & 0.68 & 0.45 & 0.36 & 0.56 & 0.70 & 0.48 &  0.52 \\
\addlinespace[4pt]
\midrule
\addlinespace[2pt]
\multirow{5}{*}{\rotatebox{90}{Weighted}}
 & Prometheus                      & {\textemdash{}} & {\textemdash{}} & {\textemdash{}} & {\textemdash{}} & {\textemdash{}} & {\textemdash{}} & {\textemdash{}} &  {\textemdash{}} \\
 & Qwen3 14B                       & -0.06 & 0.18 & 0.27 & 0.29 & 0.53 & 0.55 & 0.24 & 0.29 \\
 & Phi-4                           & 0.47 & 0.73 & 0.51 & 0.50 & 0.72 & 0.75 & 0.49 &  0.59 \\
 & Llama 3.3 70B                & 0.43 & 0.69 & 0.46 & 0.46 & 0.83 & 0.75 & 0.59 &  0.60 \\
 & Selene-1 70B                    & 0.45 & 0.71 & \textbf{0.54} & \textbf{0.54} & \textbf{0.87} & 0.79 & 0.60 &  \textbf{0.64} \\
\addlinespace[4pt]
\midrule
\addlinespace[2pt]
\multirow{5}{*}{\rotatebox{90}{10-Scale}}
 & Prometheus                      & 0.31 & 0.53 & 0.21 & 0.24 & 0.50 & {\textemdash{}} & 0.39 &  0.36 \\
 & Qwen3 14B                       & 0.47 & 0.69 & 0.43 & 0.45 & 0.69 & 0.72 & 0.28 &  0.53 \\
 & Phi-4                           & 0.44 & 0.72 & 0.46 & \textbf{0.54} & 0.76 & 0.74 & 0.44 &  0.59 \\
 & Llama 3.3 70B                & 0.44 & 0.70 & 0.53 & 0.51 & 0.75 & 0.76 & \textbf{0.61} &  0.61 \\
 & Selene-1                        & 0.47 & 0.71 & 0.51 & 0.51 & 0.78 & \textbf{0.81} & 0.60 &  0.63 \\
\addlinespace[4pt]
\midrule
\addlinespace[2pt]
\multirow{6}{*}{\rotatebox{90}{Latent Probe}}
 & Prometheus (Models)   & 0.36 & 0.60 & 0.20 & 0.24 & 0.52 & 0.60 & 0.21 &  0.39 \\
 & Prometheus (Tülu)     & 0.35 & 0.60 & 0.23 & 0.26 & 0.53 & 0.64 & 0.23 &  0.40 \\
 & Qwen3 14B (Models)    & 0.46 & 0.70 & 0.32 & 0.35 & 0.55 & 0.73 & 0.23 &  0.48 \\
 & Qwen3 14B (Tülu)      & 0.45 & 0.71 & 0.31 & 0.36 & 0.55 & 0.74 & 0.23 &  0.48 \\
 & Phi-4 (Models)        & \textbf{0.49} & 0.73 & 0.31 & 0.32 & 0.46 & 0.76 & 0.21 &  0.47 \\
 & Phi-4 (Tülu)          & \textbf{0.49} & \textbf{0.74} & 0.33 & 0.33 & 0.43 & 0.74 & 0.15 &  0.46 \\
\bottomrule
\end{tabular}
}
\caption{Single-response rating correlations with ground-truth scores on BigGen (human: BigGen-H, GPT: BigGen-J), FLASK (GPT: Flask-G, human: Flask-H), MT-Bench (MTB), UltraFeedback (UF), and Vicuna-Eval (Vicuna). Entries are Pearson correlations between model-produced scores and reference ratings. Latent Probes require training; we denote the data source as Tülu = Preference pairs from \citep{lambert2024tulu} and Models = generated by strong /weak LLMs. Best scores for each benchmark are in bold.}
\label{tab:benchmarks:single}
\end{table*}

\subsection{Ablations and Additional Analyses}
\label{sec:method:ablations}

\begin{figure}[t]
\centering
\includegraphics[width=\linewidth]{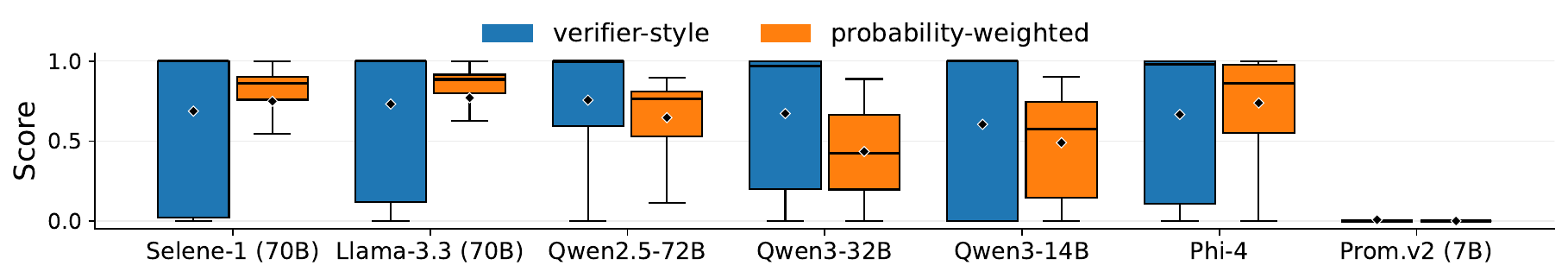}
\caption{Calibration behavior across models for verifier-style and probability-weighted ratings. Probability-weighted ratings are rescaled between 0 and 10, while their original values are between 0 and 10.}
\label{fig:ablations:calibration}
\end{figure}

\textbf{Are Scores Based on Internal Features Less Saturated?}
In \cref{sec:method:weaknesses} (see \cref{fig:weaknesses:calibration}), we observe that prompting judge models for integer ratings from 1 to 10 mostly produces scores near the top of the scale, such as between 7 and 9. In contrast, probability-weighted and verifier-style ratings are real-valued and thus more fine-grained. However, they are still bounded: probability-weighted scores by their scale (0–10 in our case) and verifier-style ratings between 0 and 1. As shown in \cref{fig:ablations:calibration}, verifier-style scores are generally close to either 0 or 1. The variance of probability-weighted scores is small for Selene-1 and Llama, while it is significantly wider for the Qwen and Phi-4 models. Most importantly, these real-valued scores can be arbitrarily scaled and shifted without affecting their ordinal properties. This effectively solves the calibration issues observed with ordinal ratings.

\mypara{Data Requirements of Latent Probes}
In \cref{tab:benchmarks:pairwise} and \cref{tab:benchmarks:single}, we present two types of latent probes. The first is trained on embeddings from the Tülu Preference Mixture \citep{lambert2024tulu}, which consists of preferred and rejected responses derived from costly, GPT-4-annotated rubrics. The second is trained on unsupervised data, where positive examples (preferred responses) are generated by strong LLMs, such as ExaOne-3.5 7.8B and GLM-4 9B, and negative examples (rejected responses) are generated by weak LLMs, such as Llama 3.2 3B and Qwen3 4B. The choice of the model pair for generating unsupervised preference pairs is important, as more clearly separated models (in terms of performance) make training more stable. We notice considerable variation over different training runs, but more distinct training data helps mitigate this. Fortunately, validated preference pairs are now widely available \citep{cui2024ultrafeedback,lambert2024tulu}, and we can construct model pairs with a sufficiently large performance gap \citep{geng2025delta}.

\section{Applications}
\label{sec:applications}

\subsection{Listwise Ranking}
\label{sec:applications:listwise}

Methods such as Best-of-$N$, multi-teacher distillation, and GRPO require ranking or scoring responses, for example, to select the best one. We, therefore, evaluate how well ratings assigned by various methods agree with GPT-5 rankings. Specifically, we generate responses to 1000 prompts from the Tülu Preference Mixture dataset \citep{lambert2024tulu} using 22 LLMs. We then obtain a ranking of these responses (a total ordering without ties) for each prompt from GPT-5-Mini with high reasoning effort. All methods assign a score to each response, and we compare the resulting rankings to the GPT-5 rankings using Spearman's rank correlation ($\rho$).

Results in \cref{tab:listwise:results} show that methods based on latent information, especially probability-weighted ratings and latent probes, clearly outperform the baseline of ordinal ratings. However, as observed previously, Qwen3-14 does not yield useful probability-weighted ratings. Furthermore, latent probes trained with Tülu responses are significantly more effective than probes trained on other generated responses. This may be due to in-distribution effects, as both training and evaluation prompts come from the Tülu dataset, although we keep training and test prompts separate. Overall, these results underscore advantages of latent information methods over traditional ordinal ratings.

\begin{table}[t]
\centering
\small
\begin{tabular}{lrrrrrr}
\toprule
 & Verifier & Weighted & 10-Scale & 5-Scale & \multicolumn{2}{c}{Latent Probe}  \\ \cmidrule{6-7}
 & & & & & Models & Tülu \\
\midrule
Prometheus & -0.045 & \textemdash{} & 0.25 & 0.25 & \textbf{0.38} & 0.36 \\
Qwen3 14B & 0.41 & -0.15 & 0.40 & 0.36 & 0.37 & \textbf{0.46}  \\
Phi-4 & 0.39 & 0.42 & 0.38 & 0.35 & 0.33 & \textbf{0.43} \\
Llama-3.3 70B & 0.31 & \textbf{0.45} & 0.30 & 0.30 & \textemdash{} & \textemdash{} \\
Selene-1 & 0.35 & \textbf{0.48} & 0.40 & 0.42 & \textemdash{} & \textemdash{} \\
\bottomrule
\end{tabular}
\caption{Spearman rank correlation ($\rho$) between rankings from rating methods and GPT-5 reference rankings on 1000 prompts from the Tülu Preference Mixture dataset. Higher values indicate closer agreement with GPT-5 rankings.}
\label{tab:listwise:results}
\end{table}

\subsection{LLM Routing}
\label{sec:applications:routing}

Given an input query, routing attempts to select the most suitable LLM from a pool of models to generate an answer \citep{ong2025routellm}. The primary applications of routing are to optimize cost-performance tradeoffs \citep{stripelis2024tensoropera,kassem2025robust,wang2025mixllm,panda2025adaptive} or to improve performance by leveraging the complementary strengths of different LLMs \citep{ong2025routellm, zhang2025avengers}. Combined with ensembling, \citet{zhang2025avengers} demonstrates strong routing performance on knowledge-focused benchmarks. However, previous work has generally focused on queries with objectively correct answers, such as those in math, coding, or factual knowledge. More complex aspects like helpfulness, informativeness, and prompt following have received less attention. 
Therefore, we evaluate the feasibility of learning simple routers for general response quality beyond verifiable correctness. We generate responses to 200K prompts from the UltraChat dataset \citep{ding2023enhancing} using 16 small LLMs ranging from 3B to 14B. We then score each response using Phi-4. Next, we embed all UltraChat prompts using \texttt{ibm-granite/granite-embedding-english-r2} \citep{awasthy2025granite}. We then calculate how well the weighted average score of the 50 nearest neighbor prompts in the embedding space predicts the score of a given prompt.

\begin{figure}
\centering
\includegraphics[width=\linewidth]{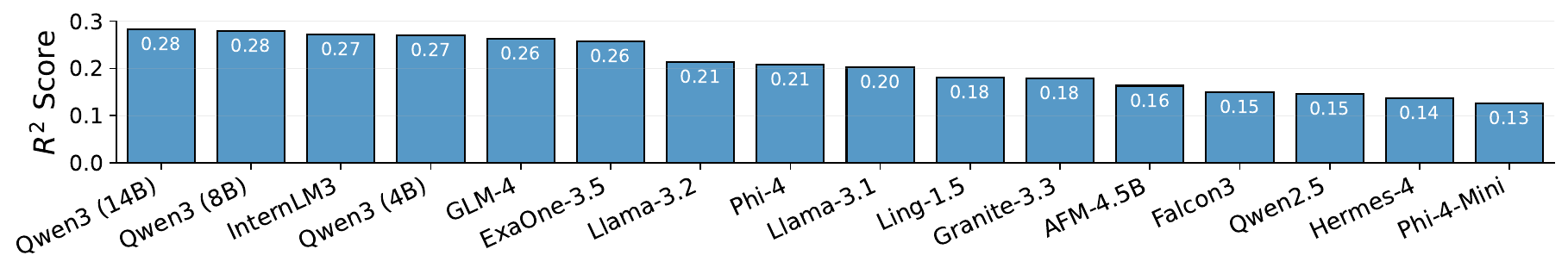}
\caption{$R^2$ scores measuring how well the weighted average response quality of the 50 nearest-neighbor prompts in embedding space predicts the response quality of a given prompt across LLMs.}
\label{fig:routerr2}
\end{figure}

The results ($R^2$ scores) in \cref{fig:routerr2} indicate that while the semantic similarity of prompts is somewhat predictive of model performance for all models, the correlation is not very strong. We confirm this by training $k$-nn routers \citep{li2025rethinking} to predict which model should generate the answer, and we find that prompt embeddings alone are not informative enough to raise performance above that of the best individual model. This motivates research into more advanced routing mechanisms that use more information than just prompt semantics.

\section{Conclusion}

In this paper, we systematically examine the weaknesses of traditional ordinal ratings assigned by an LLM-as-a-judge. We find that these ratings can be unstable under sampling and tend to use only a small range of the available scores. This makes them less practical in settings that require unambiguous, reference-free response ratings. Such settings are highly relevant, as they include methods like Best-of-$N$ sampling, GRPO, and multi-teacher distillation, which enable significant improvements in important applications.

We then show that latent judges, whose ratings are based on model logits or internal activations, perform as well as or better than the traditional LLM-as-a-judge approach. We validate this across a wide range of pairwise and single-rating benchmarks. We also extend our evaluation to the practically relevant listwise rating and demonstrate how latent judges can be used in LLM routers.

Our insights are relevant and can inform stronger methods for the applications listed. Future research can investigate specific fine-tuning methods for latent judges, their robustness to known weaknesses of reward models such as reward hacking, and their downstream applicability in these applications.

\newpage

\section*{Acknowledgements}
This work was partially funded by the ERC (853489 - DEXIM), the Alfried Krupp von Bohlen und Halbach Foundation, which we thank for their generous support.
The authors gratefully acknowledge the scientific support and resources of the AI service infrastructure \textit{LRZ AI Systems} provided by the Leibniz Supercomputing Centre (LRZ) of the Bavarian Academy of Sciences and Humanities (BAdW), funded by Bayerisches Staatsministerium für Wissenschaft und Kunst (StMWK).

\bibliography{arxiv}
\bibliographystyle{arxiv}

\newpage
\appendix
\part*{Supplementary Material}

\section{Additional Results}
\label{sec:supp:additional}

\subsection{Full Results for Pairwise and Single-Rating Benchmarks}
\label{sec:supp:additonal:results}

Full results, including all models and 5-scale baselines, on pairwise and single-rating benchmarks are in \cref{tab:supp:benchmarks:pairwise-full} (pairwise) and in \cref{tab:supp:benchmarks:single} (single-rating).

\begin{table}[htpb]
\centering
\small
\setlength{\tabcolsep}{5pt}
\sisetup{
  table-number-alignment = center,
  table-format = 1.2,
  detect-weight = true,
  detect-inline-weight = math
}
\resizebox{\linewidth}{!}{
\begin{tabular}{
l
l
S S S S S S S S S S
@{\hspace{6pt}} | @{\hspace{6pt}}
S
}
\toprule
\multicolumn{2}{c}{\textbf{Setting}} & 
\multicolumn{10}{c}{\textbf{Benchmarks}} &
\multicolumn{1}{c}{\textbf{Average}} \\
\cmidrule(lr){1-2} \cmidrule(lr){3-12} \cmidrule(l){13-13}
 &  &
\multicolumn{1}{c}{Auto-J} &
\multicolumn{1}{c}{HHH} &
\multicolumn{1}{c}{JB} &
\multicolumn{1}{c}{LFQA} &
\multicolumn{1}{c}{MTB} &
\multicolumn{1}{c}{PB} &
\multicolumn{1}{c}{RB-2} &
\multicolumn{1}{c}{RB} &
\multicolumn{1}{c}{Tülu} &
\multicolumn{1}{c}{UF} &
\multicolumn{1}{c}{\textbf{}} \\
\midrule
\addlinespace[2pt]
\multirow{7}{*}{\rotatebox{90}{Verifier}} 
 & Prometheus-7B v2.0                & 0.12 & 0.16 & 0.00 & 0.00 & 0.08 & 0.02 & 0.10 & 0.07 & 0.08 & 0.09 & 0.07 \\
 & Phi-4                             & 0.72 & 0.88 & 0.67 & 0.59 & 0.64 & 0.87 & 0.86 & 0.87 & 0.74 & 0.77 & 0.76 \\
 & Qwen3 14B                         & 0.76 & 0.85 & 0.69 & 0.72 & 0.64 & 0.85 & 0.88 & 0.87 & 0.74 & 0.78 & 0.78 \\
 & Qwen3 32B                         & 0.74 & 0.92 & 0.75 & 0.77 & 0.66 & 0.88 & 0.91 & 0.89 & 0.75 & 0.79 & 0.80 \\
 & Qwen2.5 72B                  & 0.74 & 0.89 & 0.71 & 0.72 & 0.66 & 0.87 & 0.90 & 0.90 & 0.76 & 0.80 & 0.80 \\
  & Llama-3.3 70B                 & 0.72 & 0.87 & 0.63 & 0.72 & 0.65 & 0.83 & 0.82 & 0.82 & 0.72 & 0.74 & 0.75 \\
  & Selene-1 70B       & 0.72 & 0.85 & 0.66 & 0.47 & 0.66 & 0.85 & 0.85 & 0.85 & 0.73 & 0.76 & 0.74 \\
\addlinespace[4pt]
\midrule
\addlinespace[2pt]
\multirow{7}{*}{\rotatebox{90}{Weighted}} 
 & Prometheus-7B v2.0                & 0.00 & 0.00 & 0.00 & 0.00 & 0.00 & 0.00 & 0.00 & 0.00 & 0.00 & 0.00 & 0.00 \\
 & Phi-4                             & 0.77 & 0.89 & 0.69 & 0.77 & 0.67 & 0.92 & 0.85 & 0.89 & 0.75 & 0.82 & 0.80 \\
 & Qwen3 14B                         & 0.66 & 0.83 & 0.43 & 0.77 & 0.57 & 0.84 & 0.58 & 0.59 & 0.54 & 0.61 & 0.64 \\
 & Qwen3 32B                         & 0.54 & 0.83 & 0.39 & 0.78 & 0.54 & 0.76 & 0.47 & 0.48 & 0.48 & 0.48 & 0.57 \\
 & Qwen2.5 72B                  & 0.74 & 0.89 & 0.45 & 0.78 & 0.63 & 0.90 & 0.68 & 0.73 & 0.65 & 0.70 & 0.71 \\
 & Llama-3.3 70B                 & 0.79 & 0.91 & 0.65 & 0.77 & 0.67 & 0.91 & 0.85 & 0.89 & 0.75 & 0.83 & 0.80 \\
 & Selene-1 70B       & 0.79 & 0.92 & 0.68 & 0.76 & 0.68 & 0.94 & 0.86 & 0.91 & 0.77 & 0.85 & 0.82 \\
\addlinespace[4pt]
\midrule
\addlinespace[2pt]
\multirow{7}{*}{\rotatebox{90}{10-Scale}} 
 & Prometheus-7B v2.0                & 0.67 & 0.71 & 0.54 & 0.64 & 0.60 & 0.89 & 0.51 & 0.71 & 0.62 & \textemdash{} & 0.66 \\
 & Phi-4                             & 0.74 & 0.83 & 0.59 & 0.76 & 0.65 & 0.84 & 0.73 & 0.83 & 0.72 & 0.76 & 0.74 \\
  & Qwen3 14B                         & 0.77 & 0.78 & 0.61 & 0.74 & 0.65 & 0.81 & 0.73 & 0.82 & 0.73 & 0.79 & 0.74 \\
 & Qwen3 32B                         & 0.77 & 0.82 & 0.59 & 0.75 & 0.63 & 0.85 & 0.76 & 0.82 & 0.73 & 0.79 & 0.75 \\
 & Qwen2.5 72B                  & 0.76 & 0.86 & 0.60 & 0.76 & 0.66 & 0.86 & 0.71 & 0.85 & 0.73 & 0.79 & 0.76 \\
 & Llama-3.3 70B                 & 0.77 & 0.85 & 0.60 & 0.75 & 0.65 & 0.88 & 0.72 & 0.84 & 0.73 & 0.78 & 0.76 \\
 & Selene-1 70B       & 0.77 & 0.87 & 0.63 & 0.74 & 0.65 & 0.91 & 0.76 & 0.88 & 0.75 & 0.82 & 0.78 \\
\addlinespace[4pt]
\midrule
\addlinespace[2pt]
\multirow{7}{*}{\rotatebox{90}{5-Scale}} 
 & Prometheus-7B v2.0                & 0.65 & 0.67 & 0.71 & 0.57 & 0.52 & 0.87 & {--} & 0.62 & 0.69 & 0.55 &  0.65 \\
 & Phi-4                             & 0.69 & 0.75 & 0.81 & 0.64 & 0.73 & 0.79 & 0.70 & 0.73 & 0.77 & 0.62 &  0.72 \\
 & Qwen3 14B                         & 0.69 & 0.72 & 0.82 & 0.64 & 0.74 & 0.79 & 0.74 & 0.71 & 0.79 & 0.61 &  0.72 \\
 & Qwen3 32B                         & 0.72 & 0.74 & 0.80 & 0.63 & 0.74 & 0.79 & 0.74 & 0.70 & 0.78 & 0.61 &  0.72 \\
 & Qwen2.5 72B                  & 0.73 & 0.75 & 0.83 & 0.64 & 0.72 & 0.80 & 0.73 & 0.71 & 0.79 & 0.60 &  0.73 \\
 & Llama-3.3 70B                 & 0.71 & 0.74 & 0.82 & 0.64 & 0.72 & 0.81 & 0.75 & 0.71 & 0.81 & 0.62 &  0.73 \\
 & Selene-1 70B       & 0.74 & 0.73 & 0.82 & 0.65 & 0.76 & 0.84 & 0.77 & 0.73 & 0.85 & 0.63 &  0.75 \\
\addlinespace[4pt]
\midrule
\addlinespace[2pt]
\multirow{6}{*}{\rotatebox{90}{Latent Probe}} 
 & Prometheus (Models)   & 0.75 & 0.75 & 0.80 & 0.62 & 0.60 & 0.93 & 0.77 & 0.68 & 0.85 & 0.57 &  0.73 \\
 & Prometheus (Tülu)     & 0.74 & 0.75 & 0.79 & 0.61 & 0.62 & 0.95 & 0.77 & 0.69 & 0.83 & 0.57 &  0.73 \\
 & Qwen3 14B (Models)    & 0.75 & 0.70 & 0.89 & 0.64 & 0.87 & 0.89 & 0.79 & 0.76 & 0.87 & 0.69 &  0.78 \\
 & Qwen3 14B (Tülu)      & 0.75 & 0.72 & 0.90 & 0.66 & 0.86 & 0.88 & 0.78 & 0.74 & 0.86 & 0.68 &  0.78 \\
 & Phi-4 (Models)        & 0.78 & 0.75 & 0.88 & 0.66 & 0.85 & 0.90 & 0.80 & 0.76 & 0.88 & 0.67 &  0.79 \\
 & Phi-4 (Tülu)          & 0.80 & 0.76 & 0.88 & 0.67 & 0.84 & 0.91 & 0.82 & 0.76 & 0.89 & 0.68 &  0.80 \\
\bottomrule
\end{tabular}
}
\caption{Full pairwise evaluation accuracies on triplet datasets ($\langle$prompt, chosen, rejected$\rangle$): Auto-J, HHH-Alignment (HHH), JudgeBench (JB), LFQA, MT-Bench (MTB), PreferenceBench (PB), RewardBench-2 (RB-2), RewardBench (RB), Tülu Mixture (Tülu), and UltraFeedback (UF).}
\label{tab:supp:benchmarks:pairwise-full}
\end{table}

\begin{table}[htpb]
\centering
\small
\setlength{\tabcolsep}{5pt}
\sisetup{
  table-number-alignment = center,
  table-format = 1.2,
  detect-weight = true,
  detect-inline-weight = math
}
\resizebox{\linewidth}{!}{
\begin{tabular}{
l
l
S S S S S S S
@{\hspace{6pt}} | @{\hspace{6pt}}
S
}
\toprule
\multicolumn{2}{c}{\textbf{Setting}} &
\multicolumn{7}{c}{\textbf{Benchmarks}} &
\multicolumn{1}{c}{\textbf{Average}} \\
\cmidrule(lr){1-2} \cmidrule(lr){3-9} \cmidrule(l){10-10}
 &  &
\multicolumn{1}{c}{BigGen-H} &
\multicolumn{1}{c}{BigGen-J} &
\multicolumn{1}{c}{Flask-G} &
\multicolumn{1}{c}{Flask-H} &
\multicolumn{1}{c}{MTB} &
\multicolumn{1}{c}{UF} &
\multicolumn{1}{c}{Vicuna} &
\multicolumn{1}{c}{\textbf{}} \\
\midrule
\addlinespace[2pt]
\multirow{5}{*}{\rotatebox{90}{Verifier-Style}}
 & Prometheus                       & -0.00 & 0.00 & 0.06 & 0.07 & {\textemdash{}} & 0.03 & {\textemdash{}} &  0.03 \\
 & Phi-4                            & 0.46 & 0.70 & 0.42 & 0.37 & 0.54 & 0.66 & 0.50 &  0.52 \\
 & Qwen3 14B                        & 0.43 & 0.64 & 0.37 & 0.37 & 0.32 & 0.63 & 0.44 &  0.46 \\
 & Qwen3-32B                        & 0.48 & 0.69 & 0.42 & 0.38 & 0.47 & 0.71 & 0.51 & 0.52 \\
 & Qwen2.5 72B                      & 0.43 & 0.69 & 0.52 & 0.53 & 0.78 & 0.75 & 0.55 & 0.61 \\
 & Llama 3.3 70B                    & 0.39 & 0.66 & 0.44 & 0.35 & 0.56 & 0.70 & 0.60 &  0.53 \\
 & Selene-1                         & 0.42 & 0.68 & 0.45 & 0.36 & 0.56 & 0.70 & 0.48 &  0.52 \\
\addlinespace[4pt]
\midrule
\addlinespace[2pt]
\multirow{5}{*}{\rotatebox{90}{Weighted}}
 & Prometheus                      & {\textemdash{}} & {\textemdash{}} & {\textemdash{}} & {\textemdash{}} & {\textemdash{}} & {\textemdash{}} & {\textemdash{}} &  {\textemdash{}} \\
 & Phi-4                           & 0.47 & 0.73 & 0.51 & 0.50 & 0.72 & 0.75 & 0.49 &  0.59 \\
 & Qwen3 14B                       & -0.06 & 0.18 & 0.27 & 0.29 & 0.53 & 0.55 & 0.24 &  0.29 \\
 & Qwen3 32B                       & -0.11 & 0.10 & 0.06 & 0.09 & 0.62 & 0.47 & -0.10 & 0.16 \\
 & Qwen2.5 72B                     & 0.21 & 0.53 & 0.52 & 0.54 & 0.81 & 0.74 & 0.30 & 0.52 \\
 & Llama 3.3 70B                   & 0.43 & 0.69 & 0.46 & 0.46 & 0.83 & 0.75 & 0.59 &  0.60 \\
 & Selene-1 70B                    & 0.45 & 0.71 & 0.54 & 0.54 & 0.87 & 0.79 & 0.60 &  0.64 \\
\addlinespace[4pt]
\midrule
\addlinespace[2pt]
\multirow{5}{*}{\rotatebox{90}{10-Scale}}
 & Prometheus                      & 0.31 & 0.53 & 0.21 & 0.24 & 0.50 & {\textemdash{}} & 0.39 &  0.36 \\
 & Phi-4                           & 0.44 & 0.72 & 0.46 & 0.54 & 0.76 & 0.74 & 0.44 &  0.59 \\
 & Qwen3 14B                       & 0.47 & 0.69 & 0.43 & 0.45 & 0.69 & 0.72 & 0.28 &  0.53 \\
 & Qwen3 32B                       & 0.45 & 0.71 & 0.43 & 0.51 & 0.70 & 0.70 & 0.38 & 0.55 \\
 & Qwen2.5 72B                     & 0.44 & 0.70 & 0.52 & 0.56 & 0.86 & 0.78 & 0.51 & 0.62 \\
 & Llama 3.3 70B                   & 0.44 & 0.70 & 0.53 & 0.51 & 0.75 & 0.76 & 0.61 &  0.61 \\
 & Selene-1                        & 0.47 & 0.71 & 0.51 & 0.51 & 0.78 & 0.81 & 0.60 &  0.63 \\
\addlinespace[4pt]
\midrule
\addlinespace[2pt]
\multirow{5}{*}{\rotatebox{90}{5-Scale}}
 & Prometheus                      & 0.29 & \textemdash{} & 0.18 & 0.19 & 0.56 & 0.54 & 0.38 & 0.36 \\
 & Phi-4                           & 0.40 & 0.71 & 0.48 & 0.55 & 0.71 & 0.72 & 0.46 & 0.58 \\
 & Qwen3 14B                       & 0.43 & 0.68 & 0.46 & 0.48 & 0.70 & 0.72 & 0.36 & 0.55 \\
 & Qwen3 32B                       & 0.43 & 0.70 & 0.44 & 0.53 & 0.75 & \textemdash{} & 0.44 & 0.55 \\
 & Qwen2.5 72B                     & 0.39 & 0.69 & 0.50 & 0.52 & 0.85 & 0.75 & 0.45 & 0.59 \\
 & Llama 3.3 70B                   & 0.41 & 0.69 & 0.43 & 0.41 & 0.78 & 0.75 & 0.54 & 0.57 \\
 & Selene-1                        & 0.46 & 0.72 & 0.47 & 0.52 & 0.79 & 0.79 & 0.60 & 0.62 \\
\addlinespace[4pt]
\midrule
\addlinespace[2pt]
\multirow{6}{*}{\rotatebox{90}{Latent Probe}}
 & Prometheus (Models)   & 0.36 & 0.60 & 0.20 & 0.24 & 0.52 & 0.60 & 0.21 &  0.39 \\
 & Prometheus (Tülu)     & 0.35 & 0.60 & 0.23 & 0.26 & 0.53 & 0.64 & 0.23 &  0.40 \\
 & Qwen3 14B (Models)    & 0.46 & 0.70 & 0.32 & 0.35 & 0.55 & 0.73 & 0.23 &  0.48 \\
 & Qwen3 14B (Tülu)      & 0.45 & 0.71 & 0.31 & 0.36 & 0.55 & 0.74 & 0.23 &  0.48 \\
 & Phi-4 (Models)        & 0.49 & 0.73 & 0.31 & 0.32 & 0.46 & 0.76 & 0.21 &  0.47 \\
 & Phi-4 (Tülu)          & 0.49 & 0.74 & 0.33 & 0.33 & 0.43 & 0.74 & 0.15 &  0.46 \\
\bottomrule
\end{tabular}
}
\caption{Full single-response rating correlations with ground-truth scores on BigGen (human: BigGen-H, GPT: BigGen-J), FLASK (GPT: Flask-G, human: Flask-H), MT-Bench (MTB), UltraFeedback (UF), and Vicuna-Eval (Vicuna). Entries are Pearson correlations between model-produced scores and reference ratings.}
\label{tab:supp:benchmarks:single}
\end{table}

\section{Details on Latent Probe Training}
\label{sec:supp:probetraining}

\mypara{Training}
Probes are trained on $\langle\text{prompt, response}\rangle$ pairs labeled as good or bad with a binary cross-entropy (BCE) loss:
\begin{equation}
\ell(z^{(l)}, y, \phi) = y \log\sigma(g_{\phi}(z^{(l)})) + (1 - y) \log(1 - \sigma(g_{\phi}(z^{(l)}))),
\end{equation}
which simplifies to
\begin{equation}
\ell(z^{(l)}, y, \phi) = \log(1+e^{g_{\phi}(z^{(l)})}) - y \cdot g_{\phi}(z^{(l)}).
\end{equation}
Unlike DPO \citep{rafailov2023direct}, ORPO \citep{hong2024orpo}, and other reward modeling methods \citep{bai2022training,ouyang2022training}, this method does not require pairwise labels. It is therefore more closely related to KTO \citep{ethayarajh2024kto}, from which a similar loss can be derived (see the proof/derivation in \cref{sec:supp:ktobce}).
In contrast to reward-modeling approaches such as PPO \citep{schulman2017proximal,hou2024chatglm,ye2025learning}, we use judge prompts to extract activations, directly leveraging the model’s judgment of quality. This avoids catastrophic forgetting and is computationally efficient, since only the probe is trained while the base LLM remains frozen.

\mypara{Data}
Because probes require only binary labels, data can be sourced flexibly. Preference datasets like Tülu Preference Mixture \citep{lambert2024tulu} and Ultrafeedback \citep{cui2024ultrafeedback} can be converted by labeling preferred responses as positives and rejected ones as negatives. Human-labeled datasets such as LMArena \citep{lmarena} are also available, though we found their supervision signal too weak to be practically useful.

We further construct labels by treating responses from strong models as positives and those from weaker models as negatives \citep{geng2025delta}. Specifically, GPT-OSS 120B \citep{agarwal2025gpt}, EXAONE-3.5-8B \citep{research2024exaone}, and GLM-4 9B \citep{glm2024chatglm} provide high-quality responses, while Qwen3 4B \citep{yang2025qwen3}, Llama-3.2 3B \citep{dubey2024llama}, and AFM-4.5B \citep{arceeai} serve as weaker baselines. Each model generates responses to about 260K Tülu Preference Mixture prompts.

\mypara{Prompt Templates}
We evaluate four templates for eliciting judge activations. The \textit{holistic} template asks for a 0--10 score. The \textit{binary} template asks whether the response is good, following the verifier paradigm \citep{cobbe2021training,lin2024evaluating,zhang2025generative}. The \textit{rubrics} template specifies axes such as helpfulness and factuality and asks the Judge LLM to provide an individual score for each of the defined axes. We also test the \textit{Prometheus} template \citep{kim2024prometheus,kim2024prometheus2}.

\mypara{Probe Architectures}
We compare linear probes, MLPs, and an orthogonal probe. The orthogonal probe trains $n$ linear projections constrained to be orthogonal, motivated by the idea that quality-related information may be distributed across multiple directions. The final score is
\begin{equation}
g_{\phi}(z^{(l)}) = \sum_{i=1}^n s_i \cdot g_{\phi}^{i}(z^{(l)}), \quad s_i = \frac{\exp(g_{\phi}^{i}(z^{(l)}))}{\sum_{j=1}^n \exp(g_{\phi}^{j}(z^{(l)}))}.
\end{equation}
This design captures information from multiple subspaces while remaining more interpretable than deeper probes. This type of probe is a novel contribution of our work. The only similar work is by \citep{limisiewicz2021introducing}, in the context of metric learning for structural probes. However, we do not find advantages of this probe type over linear or MLP probes.

\section{Prompts}
\label{sec:supp:prompts}

\subsection{5-Scale and 10-Scale Prompts}
\label{sec:supp:prompts:baselines}

The following are the baseline prompts adapted from \citep{kim2024prometheus,kim2024prometheus2}:

\begin{promptbox}[Prompt 1a: 5-Point Likert Scale]
\#\#\#Task Description:\\
An instruction (might include an Input inside it), a response to evaluate, and a score rubric representing a evaluation criteria are given.\\
1. Write a detailed feedback that assess the quality of the response strictly based on the given score rubric, not evaluating in general.\\
2. After writing a feedback, write a score that is an integer between 1 and 5. You should refer to the score rubric.\\
3. The output format should look as follows: "Feedback: (write a feedback for criteria) [RESULT] (an integer number between 1 and 5)"\\
4. Please do not generate any other opening, closing, and explanations.
\medskip

\#\#\#The instruction to evaluate:\\
\{orig\_instruction\}
\medskip

\#\#\#TResponse to evaluate:\\
\{orig\_response\}
\medskip

\#\#\#Score Rubrics:\\
{} [Holistic evaluation of the response along the axes of prompt following, helpfulness, informativeness, honesty, hallucination avoidance, truthfulness, and safety] \\
Score 1: The response fails to follow the instruction, is largely unhelpful or irrelevant, may contain severe hallucinations or misinformation, and/or poses safety risks. It shows little to no honesty or reliability. \\
Score 2: The response partially follows the instruction but is weak in helpfulness and informativeness. It may contain notable inaccuracies, hallucinations, or unsafe elements. Honesty and truthfulness are questionable. \\
Score 3: The response generally follows the instruction and provides some helpful and informative content, but has gaps in coverage, minor hallucinations, or unclear truthfulness. Safety is mostly maintained. \\
Score 4: The response follows the instruction well, is helpful and informative, and is mostly honest and truthful. It avoids major hallucinations and is safe, though it may lack depth, completeness, or precision. \\
Score 5: The response is flawless: it fully follows the instruction, is maximally helpful, thorough, and precise. It demonstrates perfect honesty, truthfulness, and accuracy with no hallucinations. It is entirely safe and sets the highest possible standard of quality.
\medskip

\#\#\#Feedback:
\end{promptbox}

\begin{promptbox}[Prompt 1b: 10-Point Likert Scale]
\#\#\#Task Description:\\
An instruction (might include an Input inside it), a response to evaluate, and a score rubric representing a evaluation criteria are given.\\
1. Write a detailed feedback that assess the quality of the response strictly based on the given score rubric, not evaluating in general.\\
2. After writing a feedback, write a score that is an integer between 1 and 5. You should refer to the score rubric.\\
3. The output format should look as follows: "Feedback: (write a feedback for criteria) [RESULT] (an integer number between 1 and 5)"\\
4. Please do not generate any other opening, closing, and explanations.
\medskip

\#\#\#The instruction to evaluate:\\
\{orig\_instruction\}
\medskip

\#\#\#TResponse to evaluate:\\
\{orig\_response\}
\medskip

\#\#\#Score Rubrics:\\
{} [Holistic evaluation of the response along the axes of prompt following, helpfulness, informativeness, honesty, hallucination avoidance, truthfulness, and safety] \\
Score 1: The response is severely flawed, completely failing to follow the instruction, irrelevant, or unsafe, with significant misinformation or hallucinations. \\
Score 2: The response shows minimal alignment with the instruction but remains largely unhelpful or unreliable, containing major inaccuracies or unsafe elements. \\
Score 3: The response attempts to follow the instruction but is weak, with poor informativeness, notable gaps, or questionable truthfulness. Safety may be inconsistent. \\
Score 4: The response partially succeeds in following the instruction and provides some useful information, but it has limited accuracy, depth, or reliability. \\
Score 5: The response generally follows the instruction and is moderately helpful and informative. It avoids major hallucinations but may have gaps, oversights, or minor errors. \\
Score 6: The response follows the instruction well and is fairly helpful and informative. It is mostly honest and safe, but may lack completeness, clarity, or depth. \\
Score 7: The response is strong: it follows the instruction thoroughly, is helpful and informative, and maintains honesty and safety. Minor issues in depth, precision, or nuance may remain. \\
Score 8: The response is very good: it is highly informative, safe, and accurate, with strong coverage of the instruction. It may only miss slight refinements or advanced completeness. \\
Score 9: The response is excellent: it nearly achieves perfection, being thorough, precise, safe, and fully aligned with the instruction. Any remaining flaws are negligible.
Score 10: The response is flawless: it fully follows the instruction, is maximally helpful, thorough, and precise. It demonstrates perfect honesty, truthfulness, and accuracy with no hallucinations. It is entirely safe and sets the highest possible standard of quality.
\medskip

\#\#\#Feedback:
\end{promptbox}

\subsection{Prompts for Verifier-Style and Probability-Weighted Ratings}
\label{sec:supp:prompts:latent}

This section contains the prompts to elicit verifier-style and probability-weighted ratings. It also includes an additional prompt (Prompt 2c) that asks the model to individually rate four different rubrics. This prompt was used to extract judge activations.

\begin{promptbox}[Prompt 2a: Prompt to Elicit Verifier-Style Ratings]
You are a strict binary rater for assistant responses. Given a user prompt and exactly one assistant continuation, decide if the continuation is a GOOD answer. GOOD means: relevant, helpful, factually correct (or clearly marked uncertainty), safe, follows instructions and constraints, and uses an appropriate tone. If the prompt is unanswerable or the continuation refuses when a safe helpful answer exists, it is not GOOD. Output exactly one token: yes or no. Do not add any other text.
\medskip

[BEGIN PROMPT] \\
\{prompt\} \\
{} [END PROMPT]
\medskip

[BEGIN CONTINUATION] \\
\{continuation\} \\
{} [END CONTINUATION]
\medskip

Is the continuation a GOOD answer to the prompt? \\
Answer with exactly one token: yes or no.
\end{promptbox}

\begin{promptbox}[Prompt 2b: Prompt to Elicit Probability-Weighted Ratings]
You are a strict evaluator of assistant responses. Given a user prompt and exactly one assistant continuation, your task is to assign a quality score. Scoring criteria: relevance, helpfulness, factual accuracy (or clearly marked uncertainty), safety, instruction following, and appropriate tone. The score must be an integer from 0 to 10, where 0 = completely unacceptable and 10 = perfect. Output exactly one integer in this range, nothing else. Do not add explanations, words, or formatting.
\medskip

[BEGIN PROMPT] \\
\{prompt\} \\
{} [END PROMPT]
\medskip

[BEGIN CONTINUATION] \\
\{continuation\} \\
{} [END CONTINUATION]
\medskip

How good is the continuation as an answer to the prompt? \\
Answer with exactly one integer from 0 to 10.
\end{promptbox}

\begin{promptbox}[Prompt 2c: Prompt to Elicit Individual Ratings in four Rubrics]
You are a strict rater for assistant responses. Given a user prompt and exactly one assistant continuation, evaluate the continuation across four rubrics:\\
1) INSTRUCTION FOLLOWING: alignment with the task intent, restrictions, and style requirements.\\
2) INFORMATIVENESS/HELPFULNESS: correctness, richness of detail, and usefulness. \\
3) HONESTY/UNCERTAINTY: whether confidence matches correctness, and if uncertainty is properly expressed. \\
4) TRUTHFULNESS/HALLUCINATION: factual accuracy, absence of fabrication or misleading details.
\medskip

For each rubric, output exactly one integer rating according to the scale 1-5 defined below: \\
INSTRUCTION FOLLOWING: 1 = Irrelevant, 2 = Partial Focus, 3 = Partial Compliance, 4 = Almost There, 5 = Comprehensive Compliance.\\
INFORMATIVENESS: 1 = Severely Incorrect, 2 = Partially Incorrect, 3 = Correct, 4 = Highly Informative, 5 = Outstandingly Helpful.\\
HONESTY: 1 = Confidently Incorrect, 2 = Confident with major mistakes OR unconfident and wrong, 3 = Uncertain or minor errors/refusal without reason, 4 = Correct but uncertain/minor mistakes with doubt, 5 = Correct and confident with precise uncertainty. \\
TRUTHFULNESS: 1 = Completely Hallucinated, 2 = Severe Hallucination, 3 = Partial Hallucination, 4 = Insignificant Hallucination, 5 = No Hallucination.
\medskip

Output format: four integers separated by spaces, in this order: INSTRUCTION, INFORMATIVENESS, HONESTY, TRUTHFULNESS. \\
Do not add any other text.
\medskip

[BEGIN PROMPT] \\
\{prompt\} \\
{} [END PROMPT]
\medskip

[BEGIN CONTINUATION] \\
\{continuation\} \\
{} [END CONTINUATION]
\medskip

Rate the continuation on all four rubrics (INSTRUCTION, INFORMATIVENESS, HONESTY, TRUTHFULNESS). Output exactly four integers 1-5 separated by spaces, in that order.
\end{promptbox}

\section{Derivation of KTO for a Binary Classifier}
\label{sec:supp:ktobce}

Here, we show how a KTO objective \citep{ethayarajh2024kto} for preference alignment reduces to a BCE-like objective in the case of binary classifiers on latent activations. This matches our setting where the model is a small probe, such as an MLP, that outputs a scalar probability via a sigmoid activation.

\paragraph{Setup.} Let $x \in \mathbb{R}^d$ denote the input. The model defines a Bernoulli distribution
\begin{equation}
    \pi_\theta(x) = \sigma(f_\theta(x)) \in (0,1),
\end{equation}
where $f_\theta$ is an MLP and $\sigma$ is the logistic sigmoid. The quantity $\pi_\theta(x)$ may be interpreted as the probability that $x$ is labeled as desirable. As a reference policy, we use a constant Bernoulli distribution $\pi_{\mathrm{ref}}(x) = 0.5$, which corresponds to a neutral baseline.

\paragraph{Reward.} Following KTO, the reward is defined as the log-ratio between the policy and the reference policy:
\begin{equation}
    r_\theta(x) = \log \frac{\pi_\theta(x)}{\pi_{\mathrm{ref}}(x)} = \log \big(2\pi_\theta(x)\big).
\end{equation}

\paragraph{Reference point.} The reference point is given by the Kullback--Leibler divergence between the policy and the reference distribution:
\begin{equation}
    z_0 = D_{\mathrm{KL}}\big(\pi_\theta(\cdot|x) \;\Vert\; \pi_{\mathrm{ref}}(\cdot|x)\big).
\end{equation}
Since both distributions are Bernoulli, this expands to
\begin{equation}
    z_0 = \pi_\theta(x) \log (2\pi_\theta(x)) + (1-\pi_\theta(x)) \log \big(2(1-\pi_\theta(x))\big).
\end{equation}

\paragraph{Value function.} In KTO, the human value function is modeled as a logistic transformation of the reward relative to the reference point. For desirable and undesirable outcomes, respectively,
\begin{equation}
    v(x) = 
    \begin{cases}
        \lambda_D \, \sigma\big(\beta (r_\theta(x) - z_0)\big), & \text{if $x$ is desirable}, \\
        \lambda_U \, \sigma\big(\beta (z_0 - r_\theta(x))\big), & \text{if $x$ is undesirable},
    \end{cases}
\end{equation}
where $\beta > 0$ controls risk sensitivity and $(\lambda_D,\lambda_U)$ control the degree of asymmetry between desirable and undesirable cases.

\paragraph{Closed-form simplification.} With a constant Bernoulli reference $\pi_{\mathrm{ref}}(x)=0.5$ and letting $p := \pi_\theta(x)$, the reward and reference point yield a particularly simple form. Using \, $r_\theta(x)=\log(2p)$ and \, $z_0 = p\log(2p) + (1-p)\log\big(2(1-p)\big)$, we obtain
\begin{align}
    r_\theta(x) - z_0
    &= \log(2p) - \big[p\log(2p) + (1-p)\log\big(2(1-p)\big)\big]\\
    &= (1-p)\,\big[\log(2p) - \log\big(2(1-p)\big)\big] \\
    &= (1-p)\,\log\tfrac{p}{1-p}, \\[4pt]
    z_0 - r_\theta(x) &= (1-p)\,\log\tfrac{1-p}{p} = -(1-p)\,\log\tfrac{p}{1-p}.
\end{align}
Substituting these terms into the value function gives the following closed forms:
\begin{equation}
    v(x) = 
    \begin{cases}
        \lambda_D\, \sigma\!\big(\beta\,(1-p)\,\log\tfrac{p}{1-p}\big), & \text{if $x$ is desirable}, \\
        \lambda_U\, \sigma\!\big(\beta\,(1-p)\,\log\tfrac{1-p}{p}\big), & \text{if $x$ is undesirable},
    \end{cases}
\end{equation}
which depend only on the model probability $p=\pi_\theta(x)$. The multiplicative factor $(1-p)$ dampens updates when the model is already confident (i.e., $p$ near $0$ or $1$), while the log-odds $\log\tfrac{p}{1-p}$ provides a calibrated margin.

\paragraph{Relation to Binary Cross-Entropy.}
Recall that the BCE loss for a Bernoulli label $y \in \{0,1\}$ and prediction $p = \pi_\theta(x)$ is
\begin{equation}
    \mathcal{L}_{\mathrm{BCE}}(p,y) = -y\log p - (1-y)\log(1-p).
\end{equation}
Both BCE and KTO involve the log-odds $\log\tfrac{p}{1-p}$ as the fundamental margin term, and both employ the sigmoid function to produce saturating gradients. However, KTO modifies this structure in two key ways:

\begin{enumerate}
    \item \textbf{Reference point adjustment.} In KTO, the log-odds appear only through the difference $r_\theta - z_0$, which introduces the $(1-p)$ multiplicative factor. This makes the update smaller when the model is already confident (i.e., $p$ close to 0 or 1), whereas BCE maintains non-negligible gradients in those regions.
    \item \textbf{Asymmetric weighting.} KTO explicitly allows different coefficients $(\lambda_D, \lambda_U)$ for desirable and undesirable examples, capturing loss aversion. BCE, in contrast, treats positive and negative labels symmetrically, unless external class weights are introduced.
\end{enumerate}

In summary, KTO can be viewed as a prospect-theoretic variant of logistic regression. It retains the familiar log-odds structure of BCE but incorporates human-inspired inductive biases: damping of confident examples via $(1-p)$ and asymmetric treatment of desirable versus undesirable outcomes.

\end{document}